\documentclass{article}

\usepackage{microtype}
\usepackage{graphicx}
\usepackage{booktabs} 

\usepackage{hyperref}

\usepackage[accepted]{icml2023}

\usepackage[utf8]{inputenc} 
\usepackage[T1]{fontenc}    
\usepackage{hyperref}       
\usepackage{url}            
\usepackage{booktabs}       
\usepackage{amsfonts}       
\usepackage{nicefrac}       
\usepackage{microtype}      
\usepackage{xcolor}         

\usepackage{amsmath}
\usepackage{amssymb}
\usepackage{mathtools}
\usepackage{amsthm}

\usepackage{tikz} 
\usepackage{multirow}
\usepackage{url}
\usepackage{graphicx}
\usepackage{wrapfig}
\usepackage{subcaption}
\usepackage{tcolorbox}
\usepackage{mathtools}

\usepackage[capitalize,noabbrev]{cleveref}

\def\nocheckmark{\tikz\draw[scale=0.4,fill=red,draw=red](.25,.1) -- (1,.7) -- (.25,.15) -- cycle (0.75,0.15) -- (0.77,0.15)  -- (0.3,0.7) -- cycle;}
\def\halfcheckmark{\tikz\draw[scale=0.4,fill=orange,draw=orange](0,.35) -- (.25,0) -- (1,.7) -- (.25,.15) -- cycle (0.75,0.15) -- (0.78,0.15)  -- (0.5,0.7) -- cycle;}
\def\checkmark{\tikz\draw[scale=0.4,fill=green,draw=green](0,.35) -- (.25,0) -- (1,.7) -- (.25,.15) -- cycle;}

\theoremstyle{plain}

\theoremstyle{definition}

\theoremstyle{remark}

\usepackage[textsize=tiny]{todonotes}

\icmltitlerunning{Constant Memory Attention Block}

\begin{document}

\twocolumn[
\icmltitle{Constant Memory Attention Block}



\icmlsetsymbol{equal}{*}

\begin{icmlauthorlist}
\icmlauthor{Leo Feng}{sch,comp}
\icmlauthor{Frederick Tung}{comp}
\icmlauthor{Hossein Hajimirsadeghi}{comp}
\icmlauthor{Yoshua Bengio}{sch}
\icmlauthor{Mohamed Osama Ahmed}{comp}
\end{icmlauthorlist}


\icmlaffiliation{comp}{Borealis AI}
\icmlaffiliation{sch}{Mila – Université de Montréal}

\icmlcorrespondingauthor{Leo Feng}{leo.feng@mila.quebec}

\icmlkeywords{Machine Learning, Neural Processes, Meta-Learning, Uncertainty Estimation, Efficient Computation, Attention, ICML}

\vskip 0.3in
]



\printAffiliationsAndNotice{}  

\begin{abstract}
    Modern foundation model architectures rely on attention mechanisms to effectively capture context.
    However, these methods require linear or quadratic memory in terms of the number of inputs/datapoints, limiting their applicability in low-compute domains.
    In this work, we propose Constant Memory Attention Block (CMAB), a novel general-purpose attention block that computes its output in constant memory and performs updates in constant computation. 
    Highlighting CMABs efficacy, we introduce methods for Neural Processes and Temporal Point Processes. 
    Empirically, we show our proposed methods achieve results competitive with state-of-the-art while being significantly more memory efficient.
\end{abstract}

\section{Introduction}\label{sec:intro}

The success of foundation models such as LLMs (Large Language Models) is due in no small part to the recent development of attention mechanisms.
Early attention works such as Transformers~\citep{vaswani2017attention} scaled quadratically with the number of datapoints, rendering them inapplicable to settings with large amounts of inputs.
There have been many proposed approaches to obtain efficiency gains such as sparse attention~\citep{huang2019ccnet}, low-rank self-attention~\citep{wang2020linformer}, and latent bottlenecks~\citep{goyal2021coordination, jaegle2021perceiver, lee2019set}.
For an in-depth overview, we refer the reader to the recent survey works~\citep{khan2022transformers,lin2022survey} on Transformers and their applications.
Unfortunately, these attention methods' memory requirement is at least linearly dependent (often with a large constant multiplier) on the number of inputs, limiting scalability in low compute domains (e.g., IoT devices, mobile phones and other battery-powered devices).

In this work, we propose a novel attention block called the Constant Memory Attention Block (CMAB) which allows (1) computing its output in \textbf{constant memory} regardless of the number of inputs and (2) performing updates to the attention block's input in \textbf{constant computation}.
Furthermore, CMABs do not require storing the input to update its output; as a result, unlike prior attention methods, CMABs also do not require storing the prior inputs to perform updates when deployed. 
To the best of our knowledge, we are the first to propose an attention mechanism with an efficient update mechanism that allows for computing the output of the attention block in constant memory. 
We introduce two models for different settings: Constant Memory Attentive Neural Processes (CMANPs) and Constant Memory Hawkes Process (CMHPs). 
The experimental results show these methods based on CMABs achieve results competitive with that of state-of-the-art methods while being significantly more memory efficient.

\begin{figure*}[h]
    \centering
    \includegraphics[width=0.85\textwidth]{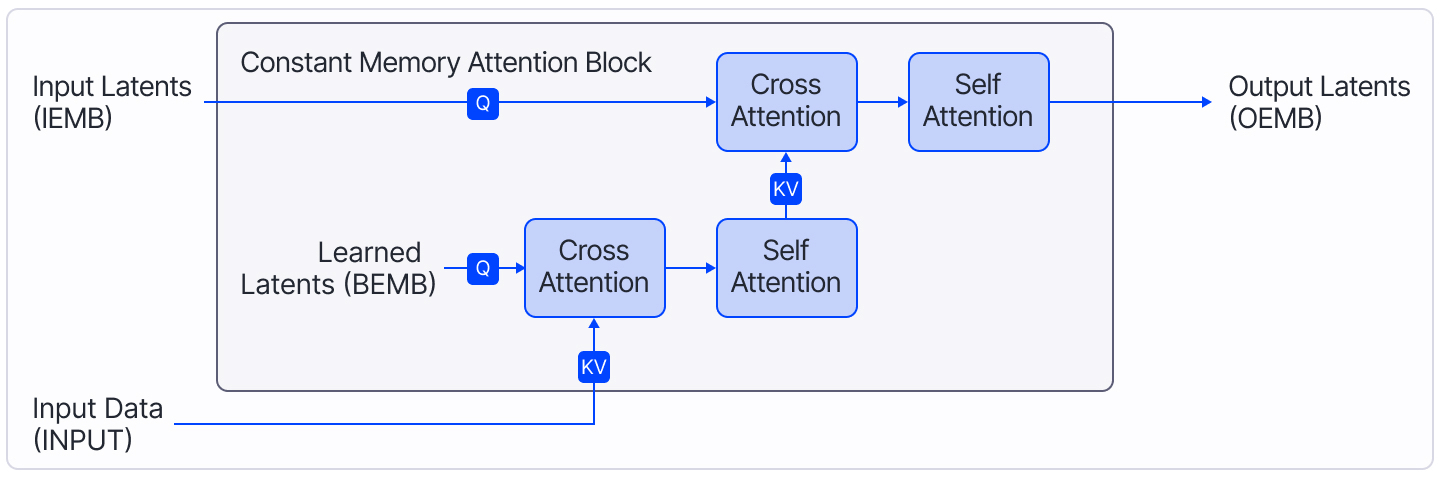}
    \caption{Constant Memory Attention Block (CMAB). 
    }
    \label{fig:euhps}
\end{figure*}
\section{Background} \label{sec:background}

\subsection{Neural Processes}

Neural Processes (NPs) are meta-learned models that efficiently compute uncertainty estimates. 
Specifically, NPs condition on an arbitrary amount of context datapoints (labelled datapoints) and make predictions for a batch of target datapoints, while preserving invariance in the ordering of the context dataset. 
Prior works have modelled NPs as $p(y | x, \mathcal{D}_\text{context}) := p(y | x, r_C)$
where $r_C := \mathrm{Agg}(\mathcal{D}_\text{context})$ such that $\mathrm{Agg}$ is a deterministic function that aggregates the context dataset $\mathcal{D}_\text{context}$ into a finite representation.
NPs are trained to maximise the likelihood of the target dataset given the context dataset.

NPs consist of three phases: conditioning, querying, and updating. 
In the conditioning phase, the model computes embeddings of the context dataset $r_C$, i.e., $r_C := \mathrm{Agg}(\mathcal{D}_\text{context})$.
During the querying phase, the model makes predictions for batches of target datapoints given the embeddings $r_C$. 
In the updating phase, the model receives new datapoints $\mathcal{D}_{update}$, and a new embedding $r'_C$  is computed, i.e., $r'_C:= \mathrm{Agg}(\mathcal{D}_{context}\,\cup\,\mathcal{D}_{update})$.

\subsection{Temporal Point Processes}
In brief, Temporal Point Processes are stochastic processes composed of a time series of discrete events. 
Recent works have proposed to model this via a neural network. 
Notably, models such as THP~\citep{zuo2020transformer} encode the history of past events to predict the next event, i.e., modelling the predictive distribution of the next event $p_\theta(\tau_{l+1} | \tau_{\leq l})$ where $\theta$ are the parameters of the model, $\tau$ represents an event, and $l$ is the number of events that have passed.
Typically, an event comprises a discrete temporal (time) stamp and a mark (categorical class). 
Models are trained to maximise the likelihood of the next event given prior events.

\section{Methodology} \label{sec:method}

\subsection{Constant Memory Attention Block (CMAB)}

The Constant Memory Attention Block (Figure \ref{fig:euhps}) takes as input the input data $\mathrm{INPUT}$ and a set of $L_I$ input latent vectors $\mathrm{IEMB}$ and outputs a set of $L_I$ output latent vectors $\mathrm{OEMB}$. 
The objective of the block is to encode the information of the input data into a fixed sized representation similar to the objective of iterative attention~\citep{jaegle2021perceiver}. 
However, unlike prior works, our proposed attention mechanism crucially allows for applying updates to the input data in constant computation per datapoint.
When stacking CMABs, the output latent vectors of the previous CMAB are fed as the input latent vectors to the next, i.e., $\mathrm{IEMB} \leftarrow \mathrm{OEMB}$. Similar to that of iterative attention, the value of $\mathrm{IEMB}$ of the first stacked CMAB block is learned. 

CMAB initially compresses the input data by applying a cross-attention between the input data and a fixed set of $L_B$ latents $\mathrm{BEMB}$ whose value is learned during training.
Next, self-attention is used to compute higher-order information:
\begin{align*}
    \mathrm{DEMB} &= \mathbf{SA}(\mathbf{CA}(\mathrm{BEMB}, \mathrm{INPUT}))
\end{align*} 
where $\mathbf{SA}$ is an abbreviation for $\mathrm{SelfAttention}$ and $\mathbf{CA}$ is an abbreviation for $\mathrm{CrossAttention}$.
Afterwards, another cross-attention between the input vectors $\mathrm{IEMB}$ and $\mathrm{DEMB}$ is performed and an additional self-attention is used to further compute higher-order information, resulting in the output vectors $\mathrm{OEMB}$:
\begin{align*}
    \mathrm{OEMB} &= \mathbf{SA}(\mathbf{CA}(\mathrm{IEMB}, \mathrm{DEMB}))
\end{align*} 

In summary, CMAB works as follows:
\begin{align*}
    \mathbf{CMAB}(\mathrm{IEMB}, \mathrm{INPUT}) &=  \\
    \mathbf{SA}(\mathbf{CA}(\mathrm{IEMB}, &\mathbf{SA}(\mathbf{CA}(\mathrm{BEMB}, \mathrm{INPUT}))))
\end{align*}
The two cross-attentions have a linear complexity of $O(N L_B)$ and a constant complexity $O(L_B L_I)$, respectively. 
The self-attentions have constant complexities of $O(L_B^2)$ and $O(L_I^2)$, respectively.
As such, the total computation required to compute the output of the block is $O(N L_B + L_B^2 + L_B L_I + L_I^2)$ where $L_B$ and $L_I$ are hyperparameter constants which bottleneck the amount of information which can be encoded.

\textbf{Constant Computation Updates.} A significant advantage of CMABs is that when given new inputs\footnote{CMABs also allow for efficient removal of datapoints (and consequently edits as well) to the input data, 
 but this is outside the scope of this work.}, CMABs can compute the updated output in constant computation per new datapoint. In contrast, a transformer block and Perceiver's iterative attention would require re-computing its output from scratch, requiring quadratic and linear computation respectively to perform a similar update.

\begin{tcolorbox}
    Having computed $\mathbf{CMAB}(\mathrm{IEMB}, \mathrm{INPUT})$ and given new datapoints $\mathcal{D}_U$ (e.g., from sequential settings such as contextual bandits or time series), $\mathbf{CMAB}(\mathrm{IEMB}, \mathrm{INPUT}\,\cup\,\mathcal{D}_U)$ can be computed in $O(|\mathcal{D}_U|)$, i.e., a constant amount of computation per new datapoint. 
\end{tcolorbox}

A formal proof and description of this process is included in the Appendix.
In brief, the proof shows that the following update procedure for the first Cross-Attention has a complexity of $O(|\mathcal{D}_U|)$:
\begin{align*}
    \mathbf{CA}(\mathrm{BEMB}, \mathrm{INPUT}\, \cup\, \mathcal{D}_U) &= \\
    \mathrm{UPD}(\mathcal{D}_U, \mathbf{CA}(\mathrm{BEMB}, &\mathrm{INPUT}))
\end{align*}
where $\mathrm{UPD}$ is an abbreviation for $\mathrm{UPDATE}$. 
Since each of the remaining self-attention and cross-attention blocks only requires constant computation. As such, CMAB can compute its updated output in $O(|\mathcal{D}_U|)$, i.e., a constant amount of computation per datapoint.

\textbf{Computing Output in Constant Memory.} 
Interestingly, a follow-up property is that CMABs can compute its output in constant memory regardless of the number of inputs.
Naively computing the output of CMAB is non-constant memory due to $\mathrm{CrossAttention}(\mathrm{BEMB}, \mathrm{CONTEXT})$ having a linear memory complexity of $O(N L_B)$. 
To achieve constant memory computation, we split the input data $\mathrm{INPUT}$ into $N/B_C$ batches of input datapoints of size up to $B_C$ (a pre-specified constant), i.e., $\mathrm{INPUT} = \cup_{i=1}^{N/B_C} \mathcal{D}_i$. Instead of computing the output at once, it is equivalent to performing the update $N/B_C - 1$ times:
\begin{align*}
    \mathbf{CA}(\mathrm{BEMB}, \mathrm{INPUT}) = \mathrm{UPD}(\mathcal{D}_{1}, 
        \mathrm{UPD}(
            \mathcal{D}_{2}, &\ldots  \\
            \mathrm{UPD}(
                \mathcal{D}_{N/B_C-1}, \mathbf{CA}(\mathrm{BEMB}, &\mathcal{D}_{N/B_C})
            )
        )
    )
\end{align*}
Computing $\mathbf{CA}(\mathrm{BEMB},\mathcal{D}_{N/B_C})$ requires $O(L_B B_C)$ constant memory. After its computation, the memory can be freed up, so that each of the subsequent $\mathrm{UPDATE}$ operations can re-use the memory space. Each of the update operations cost $O(L_B B_C)$ constant memory, resulting in $\mathbf{CA}(\mathrm{BEMB}, \mathrm{INPUT})$ only needing constant memory $O(L_B B_C)$. As a result, CMAB's output can be computed in constant memory. 

CMABs are generally useful in that they are a more memory efficient alternative to transformer or iterative attention in many settings. 
Another advantage of CMABs over prior works is that the input data does not need to be stored when performing updates with new data, meaning the model has privacy-preserving properties and is applicable to streaming data settings (e.g., settings where the data is not stored).

\subsection{Constant Memory Attentive Neural Processes (CMANP)}

In this section, we leverage CMABs to construct an efficient Neural Process by replacing the iterative attention blocks used in LBANPs~\citep{feng2023latent} with CMABs.
CMANPs (Figure \ref{fig:CMANPs} in Appendix due to space limitations) comprise of stacked CMAB blocks which take as input the context dataset. 
In Table \ref{table:memory_complexities}, we compare the memory complexities of state-of-the-art Neural Processes with that of CMANPs, showcasing the efficiency gains of CMANPs over prior methods. 
Full details regarding the computation of the conditioning, querying, and updating phase of CMANPs is included in the Appendix.

\begin{table}[]
\centering
\begin{tabular}{|l|ccccc|}
\hline
                     & \multicolumn{5}{c|}{\textbf{Memory Complexity}}                                                                                                                       \\ \cline{2-6}
                     & \multicolumn{1}{c|}{\textbf{Condition}} & \multicolumn{2}{c|}{\textbf{Query}}                              & \multicolumn{2}{c|}{\textbf{Update}}                     \\ \hline
\textbf{In Terms of} & \multicolumn{1}{c|}{$N$}                & \multicolumn{1}{c|}{$N$} & \multicolumn{1}{c|}{$M$}              & \multicolumn{1}{c|}{$N$}              & $|\mathcal{D}_U|$            \\  \hline
\textbf{TNP-D}       & \multicolumn{1}{c|}{N/A}                & \multicolumn{1}{c|}{\nocheckmark}    & \multicolumn{1}{c|}{\nocheckmark}                 & \multicolumn{1}{c|}{N/A}              & N/A              \\ 
\textbf{EQTNP}       & \multicolumn{1}{c|}{\nocheckmark}   & \multicolumn{1}{c|}{\halfcheckmark}   & \multicolumn{1}{c|}{\halfcheckmark} & \multicolumn{1}{c|}{\nocheckmark} & \nocheckmark \\ 
\textbf{LBANP}       & \multicolumn{1}{c|}{\halfcheckmark}   & \multicolumn{1}{c|}{\checkmark}   & \multicolumn{1}{c|}{\halfcheckmark} & \multicolumn{1}{c|}{\halfcheckmark} & \halfcheckmark \\ \hline
\textbf{CMANP}      & \multicolumn{1}{c|}{\checkmark}                  & \multicolumn{1}{c|}{\checkmark}   & \multicolumn{1}{c|}{\halfcheckmark}                & \multicolumn{1}{c|}{\checkmark}                & \checkmark \\  \hline
\end{tabular}
\caption{
    Comparison of Memory Complexities of state-of-the-art Neural Processes with respect to the number of context datapoints $N$, number of target datapoints in a batch $M$, and a set of new datapoints in an update $\mathcal{D}_U$.
    (Green) Checkmarks represent constant memory, (Orange) half checkmarks represent linear memory, and (Red) Xs represent quadratic or more memory. 
}
\label{table:memory_complexities}
\vspace{-5mm}
\end{table}

In leveraging CMABs, CMANP do not require the context dataset when updating the model, allowing for streaming data settings such as bayesian optimization and contextual bandit settings. 
Unlike prior work NP, the raw data would not need to be stored which is a significant advantage in settings with limited resources or where data privacy is a concern. 
In addition, CMANPs only require constant memory to perform the conditioning, querying, and updating phases of Neural Proesses, making a state-of-the-art NP highly accessible for small devices.

\subsection{Constant Memory Hawkes Processes (CMHPs)}

Building on CMABs, we introduce the Constant Memory Hawkes Process (CMHPs) (Figure \ref{appendix:fig:cmhps} in Appendix due to space limitations), a model which replaced the transformer layers in Transformer Hawkes Process (THP)~\citep{zuo2020transformer} with Constant Memory Attention Blocks. 
However, unlike THPs which summarise the information for prediction in a single vector, CMHPs summarise it into a set of latent vectors. 
As such, a flatten operation is added at the end of the model.
Following prior work~\citep{bae2023meta,Shchur2020Intensity-Free}, we use a mixture of log-normal distribution as the decoder for both THP and CMHP.
Crucially, when deployed, CMHPs do not need to store any of its history of events to update the model with new events that happen. 
Furthermore, CMHPs also only use constant memory instead of THP's quadratic memory requirement, making it a reliable model for low-memory devices.

\begin{table*}[h]
\centering
\begin{tabular}{|c|ccc|cc|}
\hline
\multirow{2}{*}{Method}      & \multicolumn{3}{c|}{CelebA}                & \multicolumn{2}{c|}{EMNIST}  \\
                             & 32x32        & 64x64        & 128x128      & Seen (0-9)  & Unseen (10-46) \\ \hline
CNP~\citep{garnelo2018conditional}                          & 2.15 ± 0.01  & 2.43 ± 0.00  & 2.55 ± 0.02  & 0.73 ± 0.00 & 0.49 ± 0.01    \\
CANP~\citep{kim2019attentive}                         & 2.66 ± 0.01  & 3.15 ± 0.00  & ---          & 0.94 ± 0.01 & 0.82 ± 0.01    \\
NP~\citep{garnelo2018neural}                           & 2.48 ± 0.02  & 2.60 ± 0.01  & 2.67 ± 0.01  & 0.79 ± 0.01 & 0.59 ± 0.01    \\
ANP~\citep{kim2019attentive}                          & 2.90 ± 0.00  & ---          & ---          & 0.98 ± 0.00 & 0.89 ± 0.00    \\
BNP~\citep{lee2020bootstrapping}                          & 2.76 ± 0.01  & 2.97 ± 0.00  & ---          & 0.88 ± 0.01 & 0.73 ± 0.01    \\
BANP~\citep{lee2020bootstrapping}                         & 3.09 ± 0.00  & ---          & ---          & 1.01 ± 0.00 & 0.94 ± 0.00    \\
TNP-D~\citep{zuo2020transformer}                        & 3.89 ± 0.01  & 5.41 ± 0.01  & ---          & 1.46 ± 0.01 & 1.31 ± 0.00    \\
LBANP~\citep{feng2023latent}                & 3.97 ± 0.02  & 5.09 ± 0.02  & 5.84 ± 0.01  & 1.39 ± 0.01 & 1.17 ± 0.01    \\ \hline
CMANP (Ours)     & 3.93 ± 0.05 & 5.02 ± 0.14 & 5.55 ± 0.01 & 1.36 ± 0.01         & 1.09 ± 0.01            \\ \hline
\end{tabular}
\caption{Image Completion Experiments. Each method is evaluated with 5 different seeds according to the log-likelihood (higher is better). The "dash" represents methods that could not be run because of the large memory requirement.}
\label{table:image_completion}
\end{table*}

\begin{table*}[]
\centering
\begin{tabular}{|c|ccc|ccc|}
\hline
\multirow{2}{*}{Method} & \multicolumn{3}{c|}{Mooc}                                                                                              & \multicolumn{3}{c|}{Reddit}                                                                                           \\ 
\multicolumn{1}{|c|}{} & \multicolumn{1}{c|}{RMSE}                    & \multicolumn{1}{c|}{NLL}                      & ACC                     & \multicolumn{1}{c|}{RMSE}                    & \multicolumn{1}{c|}{NLL}                     & ACC                     \\ \hline
THP                   & \multicolumn{1}{c|}{0.202 ± 0.017}          & \multicolumn{1}{c|}{0.267 ± 0.164}  & 0.336 ± 0.007 & \multicolumn{1}{c|}{0.238 ± 0.028} & \multicolumn{1}{c|}{0.268 ± 0.098} & 0.610 ± 0.002 \\ \hline
CMHP (Ours)                 & \multicolumn{1}{c|}{0.168 ± 0.011} & \multicolumn{1}{c|}{-0.040 ± 0.620} & 0.237 ± 0.024          & \multicolumn{1}{c|}{0.262 ± 0.037} & \multicolumn{1}{c|}{0.528 ± 0.209} & 0.609 ± 0.003 \\ \hline
\end{tabular}
\caption{Temporal Point Processes Experiments.}
\label{appendix:table:tpp}
\end{table*}

\section{Experiments} \label{sec:experiments}

\subsection{CMANPs: Image Completion}

In this experiment, we compare CMANPs against prior NP methods on standard NP datasets: EMNIST~\citep{cohen2017emnist} and CelebA~\citep{liu2015faceattributes}. TNP-D (Transformer-based model) and LBANP (Perceiver's iterative attention-based model) are the state-of-the-art for comparison.

\textbf{Results.} In Table \ref{table:image_completion}, we compare CMANPs with existing NP baselines, showing their performance is competitive with that of prior state-of-the-art: TNP-D and LBANP. 
Although all baseline methods were able to be evaluated on CelebA (32 x 32) and EMNIST, many were not able to scale to CelebA (128 x 128) due to their memory cost, including TNP-D. 
In contrast, CMANP was not affected by this limitation due to only requiring constant memory.

\subsection{CMHPs: Temporal Point Processes}

In this experiment, we compare CMHPs (CMAB-based model) against THPs (Transformer-based model) on standard TPP datasets: Mooc and Reddit (dataset details in Appendix). 
The results (Table \ref{appendix:table:tpp}) show CMHPs are competitive with THPs. Crucially, unlike THP, CMHP has the ability to efficiently update the model with new data as it arrives overtime which is typical in time series data such as in Temporal Point Processes. CMHP only requires constant computation to perform the update unlike the quadratic computation required by THP.

\subsection{Analysis}

\begin{figure}\centering
     \centering
     \includegraphics[width=0.85\linewidth]{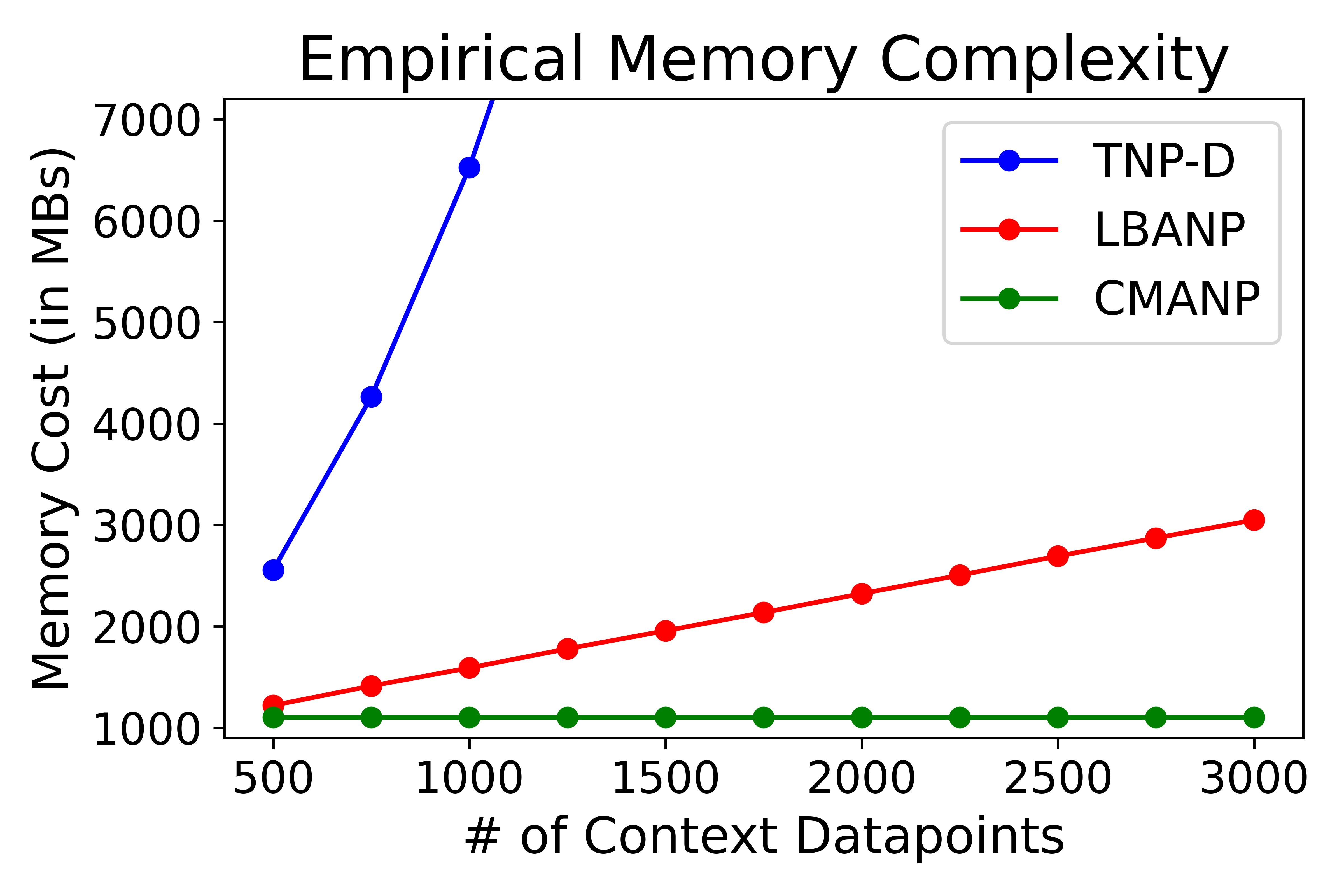}
     \hfill
    \caption{Memory Analyses Graphs.}
    \label{fig:memory_analyses}
\end{figure}

\textbf{Empirical Memory:} In Figure \ref{fig:memory_analyses}, we compare CMANP's empirical memory cost with that of state-of-the-art NP methods during evaluation. Comparing the vanilla variants of NPs, we see that TNP-D (Transformer-based model) and LBANP (Perceiver's iterative attention-based model) scale quadratically and linearly respectively with respect to the number of context datapoints. In contrast, CMANPs are significantly more efficient only requiring a low constant amount of memory. 

\section{Conclusion} \label{sec:conclusion}

In this work, we introduced CMAB (Constant Memory Attention Block), a novel efficient attention block  capable of computing its output in \textbf{constant} additional memory.
Building on CMAB, we proposed Constant Memory Attentive Neural Processes (CMANPs) and Constant Memory Hawkes Processes (CMHPs). 
Our experiments show CMANPs and CMHPs achieve results competitive with state-of-the-art while only requiring constant memory, making it applicable to settings such as low-memory domains. 
In contrast, prior state-of-the-art method required memory that is linear or quadratic in the number of datapoints.

\newpage

\bibliography{example_paper}
\bibliographystyle{icml2023}


\newpage
\appendix

\section{Appendix: Model and Experiment Details}

\begin{figure*}[h]
    \centering
    \includegraphics[width=0.75\textwidth]{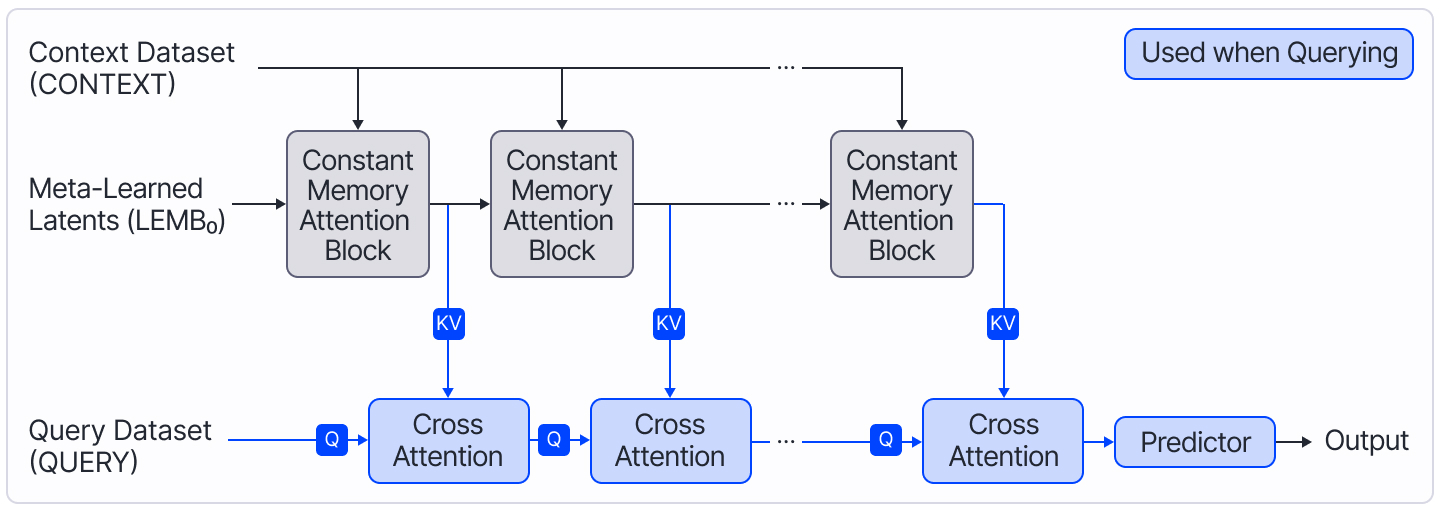}
    \caption{Constant Memory Attentive Neural Processes}
    \label{fig:CMANPs}
\end{figure*}

\begin{figure*}[h]
    \centering
    \includegraphics[width=0.8\textwidth]{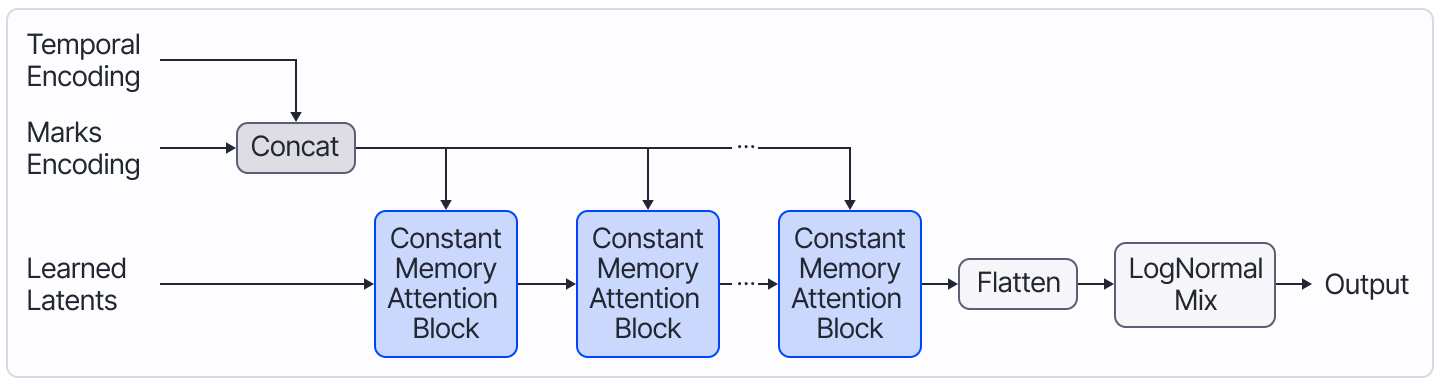}
    \caption{Constant Memory Hawkes Processes}
    \label{appendix:fig:cmhps}
\end{figure*}

\subsection{CMANPs: Conditioning, Querying, and Updating Phases}

The conditioning, querying, and updating phases in CMANPs work as follows:

\textbf{Conditioning Phase:} In the conditioning phase, the CMAB blocks encode the context dataset into a constant number of latent vectors $\mathrm{LEMB}_i$.
The first block takes as input a set of meta-learned latent vectors $\mathrm{LEMB}_0$ (i.e., $\mathrm{IEMB}$ in CMABs) and the context dataset $\mathrm{CONTEXT}$ and outputs a set of encodings $\mathrm{LEMB}_1$ (i.e., $\mathrm{OEMB}$ in CMABs).  
The output latents of each block are passed as the input latents to the next CMAB block. 
\begin{align*}
    \mathrm{LEMB}_i &= \mathbf{CMAB}(\mathrm{LEMB}_{i-1}, \mathrm{CONTEXT})
\end{align*}
Since CMAB can compute its output in constant memory, thus CMANPs can also perform this conditioning phase in constant memory. 

\textbf{Querying Phase:} In the querying phase, the deployed model retrieves information from the fixed size outputs of the CMAB blocks ($\mathrm{LEMB}_i$) to make predictions for the query dataset (QUERY). 
When making a prediction for query datapoints, information is retrieved via cross-attention. 
\begin{align*}
    \mathrm{QEMB}_0 &= \mathrm{QUERY} \\ 
    \mathrm{QEMB}_i &= \mathrm{CrossAttention}(\mathrm{QEMB}_{i-1}, \mathrm{LEMB}_{i}) \\
    \mathrm{Output} &= \mathrm{Predictor}(\mathrm{QEMB_K})
\end{align*}

\textbf{Update Phase:} In the update phase, the NP receives a batch of new datapoints $\mathcal{D}_U$ to include in the context dataset. 
CMANPs leverage the efficient update mechanism of CMABs to achieve efficient updates (constant per datapoint) to its context dataset. 
Specifically, the first CMAB block updates its output using the new datapoints. 
Afterwards, the next CMAB blocks are updated sequentially using the updated output of the previous CMAB block as follows:
\begin{align*}
    \mathrm{LEMB}'_0 &= \mathrm{LEMB}_0 \\ 
    \mathrm{LEMB}'_i &= \mathbf{CMAB}(\mathrm{LEMB}'_{i-1}, \mathrm{CONTEXT}\,\cup\,\mathcal{D}_U)
\end{align*}
Since CMAB can compute the output and perform updates in constant memory irrespective of the number of context datapoints, CMANPs can also compute its output and perform the update in constant memory.

\subsection{CMANPs: Additional Experiment Details}
We compare CMANPs against the large variety of members of the Neural Process family: Conditional Neural Processes (CNPs)~\citep{garnelo2018conditional}, Neural Processes (NPs)~\citep{garnelo2018neural}, Bootstrapping Neural Processes (BNPs)~\citep{lee2020bootstrapping}, (Conditional) Attentive Neural Processes (C)ANPs~\citep{kim2019attentive}, and Bootstrapping Attentive Neural Processes (BANPs)~\citep{lee2020bootstrapping}.
In addition, we compare against the recent state-of-the-art methods: Latent Bottlenecked Attentive Neural Processes (LBANPs)~\citep{feng2023latent} and Transformer Neural Processes (TNPs)~\citep{nguyen2022transformer}.

For the purpose of consistency, we set the number of latents (i.e., bottleneck size) $L_I = L_B = 128$ across all experiments. Similarly, for LBANPs, we report results with the number of latents (i.e., bottleneck size) $L = 128$ across all experiments.

\subsubsection{CMANPs: Image Classification Details}
The model is given a set of pixel values of an image and aims to predict the remaining pixels of the image. Each image corresponds to a unique function~\citep{garnelo2018neural}. In this experiment, the $x$ values are rescaled to [-1, 1] and the $y$ values are rescaled to $[-0.5, 0.5]$. For each task, a randomly selected set of pixels are selected as context datapoints and target datapoints. 

EMNIST comprises of black and white images of handwritten letters of $32 \times 32$ resolution. $10$ classes are used for training. The context and target datapoints are sampled according to $N \sim \mathcal{U}[3, 197)$ and $M \sim \mathcal{U}[3, 200-N)$  respectively.
CelebA comprises of colored images of celebrity faces. Methods are evaluated on various resolutions to show scalability of the methods. 
In CelebA32, images are downsampled to $32 \times 32$ and the number of context and target datapoints are sampled according to $N \sim \mathcal{U}[3, 197)$ and $M \sim \mathcal{U}[3, 200-N)$ respecitvely. 
In CelebA64, the images are down-sampled to $64 \times 64$  and $N \sim \mathcal{U}[3, 797)$ and $M \sim \mathcal{U}[3, 800-N)$.
In CelebA128, the images are down-sampled to $128 \times 128$ and $N \sim \mathcal{U}[3, 1597)$ and $M \sim \mathcal{U}[3, 1600-N)$.

\subsubsection{CMHPs: TPP Dataset Details}
\textbf{Mooc Dataset} comprises of $7,047$ sequences. Each sequence contains the action times of an individual user of an online Mooc course with 98 categories for the marks.

\textbf{Reddit Dataset} comprises of $10,000$ sequences. Each sequence contains the action times from the most active users with marks being one of the $984$ the sub-reddit categories of each sequence.

\subsection{Reproducibility}
We use the implementation of the baselines from the official repository of TNPs (\href{https://github.com/tung-nd/TNP-pytorch}{https://github.com/tung-nd/TNP-pytorch}) and LBANPs (\href{https://github.com/BorealisAI/latent-bottlenecked-anp}{https://github.com/BorealisAI/latent-bottlenecked-anp}). The datasets are standard for Neural Processes and are available in the same link.
Details regarding the architecture and the implementation is included in the main paper.
Additional details regarding the hyperparameters and architecture are included in the Appendix.

\subsection{Implementation and Hyperparameter Details}
We follow closely the hyperparameters of TNPs and LBANPs.
In CMANPs, the number of blocks for the conditioning phase is equivalent to the number of blocks in the conditioning phase of LBANPs.
Similarly, the number of cross-attention blocks for the querying phase is equivalent to that of LBANPs.
We used an ADAM optimizer with a standard learning rate of $5e-4$. 
We performed a grid-search over the weight decay term $\{0.0, 0.00001, 0.0001, 0.001\}$. 
Consistent with prior work~\citep{feng2023latent} who set their number of latents $L=128$, we also set the number of latents to the same fixed value $L_I = L_B = 128$ without tuning. 
Due to CMANPs and CMABs architecture, they allow for varying embedding sizes for the learned latent values ($\mathrm{LEMB_0}$ and $\mathrm{BEMB}$). 
For simplicity, we set the embedding sizes to $64$ consistent with prior works~\citep{nguyen2022transformer,feng2023latent}. 
During training, CelebA (128x128), (64x64), and (32x32) used a mini-batch size of 25, 50, and 100 respectively. All experiments are run with $5$ seeds.
All experiments were either run on a GTX
1080ti (12 GB RAM) or P100 GPU (16 GB RAM).

\section{Appendix: Proofs}

\subsection{CMAB's Constant Computation Updates Proof}

\textbf{Proof Outline:} Since $L_B$ and $L_I$ are constants (hyperparameters), CMAB's complexity is constant except for the contributing complexity part of the first attention block: $\mathrm{CrossAttention}(\mathrm{BEMB}, \mathrm{INPUT})$, which has a complexity of $O(N L_B)$.
As such, to achieve constant computation updates, it suffices that the updated output of this cross-attention can be updated in constant computation per datapoint. Simplified, $\mathrm{CrossAttention}(\mathrm{BEMB}, \mathrm{INPUT})$ is computed as follows:
$$\mathrm{CrossAttention}(\mathrm{BEMB}, \mathrm{INPUT}) = \mathrm{softmax}(Q K^T)V
$$
where $K$ and $V$ are key, value matrices respectively that represent the embeddings of $\mathrm{INPUT}$ and $Q$ is the query matrix representing the embeddings of $\mathrm{BEMB}$.
When an update with $\mathcal{D}_U$ new datapoints occurs, $|\mathcal{D}_U|$ rows are added to the key, value matrices. However, the query matrix is constant due to $\mathrm{BEMB}$ being a fixed set of latent vectors whose values are learned. As a result, the output of the cross-attention can be computed via a rolling average in $O(|\mathcal{D}_U|)$. 

\textbf{Full Proof:}
Recall, CMAB works as follows:
\begin{align*}
    \mathbf{CMAB}(\mathrm{IEMB}, \mathrm{INPUT}) &= \\
    \mathbf{SA}(\mathbf{CA}(\mathrm{IEMB}, \mathbf{SA}&(\mathbf{CA}(\mathrm{BEMB}, \mathrm{INPUT}))))
\end{align*}
where $\mathbf{SA}$ represents $\mathrm{SelfAttention}$ and $\mathbf{CA}$ represents $\mathrm{CrossAttention}$.
The two cross-attentions have a linear complexity of $O(N L_B)$ and a constant complexity $O(L_B L_I)$, respectively. 
The self-attentions have constant complexities of $O(L_B^2)$ and $O(L_I^2)$, respectively.
As such, the total computation required to compute the output of the block is $O(N L_B + L_B^2 + L_B L_I + L_I^2)$ where $L_B$ and $L_I$ are hyperparameter constants which bottleneck the amount of information which can be encoded.

Importantly, since $L_B$ and $L_I$ are constants (hyperparameters), CMAB's complexity is constant except for the contributing complexity part of the first attention block: $\mathrm{CrossAttention}(\mathrm{BEMB}, \mathrm{INPUT})$, which has a complexity of $O(N L_B)$.
To achieve constant computation updates, it suffices that the updated output of this cross-attention can be updated in constant computation per datapoint. Simplified, $\mathrm{CrossAttention}(\mathrm{BEMB}, \mathrm{INPUT})$ is computed as follows:
\begin{align*}
    \mathrm{emb} = \mathrm{CrossAttention}(\mathrm{BEMB}, \mathrm{INPUT}) &= \\
    \mathrm{softmax}&(Q K^T)V
\end{align*}
where $K$ and $V$ are key, value matrices respectively that represent the embeddings of $\mathrm{INPUT}$ (sets of $N$ vectors) and $Q$ is the query matrix representing the embeddings of $\mathrm{BEMB}$ (a set of $L_B$ vectors).
When an update with $\mathcal{D}_U$ new datapoints occurs, $|\mathcal{D}_U|$ rows are added to the key, value matrices. However, the query matrix is constant due to $\mathrm{BEMB}$ being a fixed set of latent vectors whose values are learned. 

Without loss of generality, for simplicity, we consider the $j-th$ output vector of the cross-attention ($\mathrm{emb}_j$). 
Let $s_i = Q_{j,:}(K_{i,:})^T$ and $v_i = V_{i, :}$, then we have the following:
$$
    \mathrm{emb}_j = \sum_{i=1}^{N} \frac{\exp(s_i)}{C} v_i
$$
where $C = \sum_{i=1}^N \exp(s_i)$. Performing an update with a set of new inputs $D_U$, results in adding $|\mathcal{D}_U|$ rows to the $K, V$ matrices and the following:

$$
    \mathrm{emb'}_j = \sum_{i=1}^{N+|\mathcal{D}_U|} \frac{\exp(s_i)}{C'} v_i
$$
where $C' = \sum_{i=1}^{N+|\mathcal{D}_U|} \exp(s_i) = C + \sum_{i=N+1}^{N+|\mathcal{D}_U|} \exp(s_i)$. As such, the updated embedding $\mathrm{emb'}_j$ can be computed via a rolling average:
$$
    \mathrm{emb'_j} = \frac{C}{C'} \times \mathrm{emb_j} \,\,+ \sum_{i=N+1}^{N+|\mathcal{D}_U|} \frac{e^{s_i}}{C'} v_i
$$
Computing $\mathrm{emb'}_j$ and $C'$ via this rolling average only requires $O(|\mathcal{D}_U|)$ operations when given $C$ and $\mathrm{emb}$ as required.
In practice, however, this is not stable. The computation can quickly run into numerical issues such as overflow problems.

\textbf{Practical Implementation:} In practice, instead of computing and storing $C$ and $C'$, we instead compute and store $\log (C)$ and $\log(C')$. 

The update is instead computed as follows: $\log(C') = \log(C) + \mathrm{softplus}(T)$ where $T = \log(\sum_{i=N+1}^{N+|\mathcal{D}_U|} \exp(s_i - \log(C)))$. $T$ can be computed efficiently and accurately using the log-sum-exp trick in $O(|\mathcal{D}_U|)$. This results in an update as follows:
\begin{align*}
    \mathrm{emb'_j} &= \exp(\log(C) - \log(C')) \times \mathrm{emb_j} + & \\
    & \sum_{i=N+1}^{N+|\mathcal{D}_U|} \exp(s_i - \log(C')) v_i
\end{align*}
The resulting $emb'$ and $C'$ is the same. However, this method of implementation avoids the numerical issues that will occur.

\textbf{Practical Implementation (Proof): } 

$$
    C = \sum_{i=1}^{N} \exp(s_i) \quad\quad C' = \sum_{i=1}^{N+|\mathcal{D}_U|} \exp(s_i)
$$

\begin{align*}
    \log(C') - \log(C) =& \\
    \log(\sum_{i=1}^{N+|\mathcal{D}_U|} &\exp(s_i)) - \log(\sum_{i=1}^{N} \exp(s_i)) \\
\end{align*}
\begin{align*}
    \log(C') &= \log(C) + \log(\frac{\sum_{i=1}^{N+|\mathcal{D}_U|} \exp(s_i)}{\sum_{i=1}^{N} \exp(s_i)}) \\
    \log(C') &= \log(C) + \log(1 + \frac{\sum_{i=N+1}^{N+|\mathcal{D}_U|} \exp(s_i)}{\sum_{i=1}^{N} \exp(s_i)}) \\
    \log(C') &= \log(C) + \log(1 + \frac{\sum_{i=N+1}^{N+|\mathcal{D}_U|} \exp(s_i)}{\exp(\log(C))}) \\
    \log(C') &= \log(C) + \log(1 + \sum_{i=N+1}^{N+|\mathcal{D}_U|} \exp(s_i - \log(C))) \\
\end{align*}
Let $T = \log(\sum_{i=N+1}^{N+|\mathcal{D}_U|} \exp(s_i - \log(C)))$. Note that $T$ can be computed efficiently using the log-sum-exp trick in $O(|\mathcal{D}_U|)$. Also, recall the $\mathrm{softplus}$ function is defined as follows: $\mathrm{softplus}(T) = \log(1+ \exp(T ))$. As such, we have the following:

\begin{align*}
    \log(C') &= \log(C) + \log(1 + \exp(T)) \\
    &= \log(C) + \mathrm{softplus}(T)
\end{align*}
Recall:
$$
    \mathrm{emb'_j} = \frac{C}{C'} \times \mathrm{emb_j} \,\,+ \sum_{i=N+1}^{N+|\mathcal{D}_U|} \frac{\exp(s_i)}{C'} v_i
$$
Re-formulating it using $\log(C)$ and $\log(C')$ instead of $C$ and $C'$ we have the following update:
\begin{align*}
    \mathrm{emb'_j} = \exp(\log(C) - &\log(C')) \times \mathrm{emb_j} +  \\ 
    &\sum_{i=N+1}^{N+|\mathcal{D}_U|} \exp(s_i - \log(C')) v_i
\end{align*}

 which only requires $O(|\mathcal{D}_U|)$ computation (i.e., constant computation per datapoint) while avoiding numerical issues. 

\subsection{Additional Properties}

In this section, we show that CMANPs uphold the context and target invariance properties. 

\textbf{Property: Context Invariance.} A Neural Process $p_\theta$ is context invariant if for any choice of permutation function $\pi$, context datapoints $\{(x_i, y_i)\}_{i=1}^{N}$, and target datapoints $x_{N+1:N+M}$, 
\begin{align*}
p_\theta(y_{N+1:N+M} | x_{N+1:N+M} , x_{1:N}, y_{1:N}) = \\
p_\theta(y_{N+1:N+M} | x_{N+1:N+M} , x_{\pi(1):\pi(N)}
, y_{\pi(1):\pi(N)})
\end{align*}

\textbf{Proof Outline:} Since CMANPs retrieve information from a compressed encoding of the context dataset computed by CMAB (Constant Memory Attention Block). It suffices to show that CMABs compute their output while being order invariant in their input (i.e., context dataset in CMANPs) ($\mathrm{INPUT}$). 

Recall CMAB's work as follows:
\begin{align*}
    \mathbf{CMAB}(\mathrm{IEMB}, \mathrm{INPUT}) &= \\
    \mathbf{SA}(\mathbf{CA}(\mathrm{IEMB}, &\mathbf{SA}(\mathbf{CA}(\mathrm{BEMB}, \mathrm{INPUT}))))
\end{align*}

where $\mathrm{IEMB}$ are a set of vectors outputted by prior blocks, $\mathrm{BEMB}$ are a set of vectors whose values are learned during training, and $\mathrm{INPUT}$ are the set of inputs in which we wish to be order invariant in.

The first cross-attention to be computed is: $\mathbf{CA}(\mathrm{BEMB}, \mathrm{INPUT})$. A nice feature of cross-attention is that its order invariant in the keys and values; in this case, these are embeddings of $\mathrm{INPUT}$. In other words, the output of $\mathbf{CA}(\mathrm{BEMB}, \mathrm{INPUT})$ is order invariant in the input data $\mathrm{INPUT}$. 

Since the remaining self-attention and cross-attention blocks take as input: $\mathrm{IEMB}$ and the output of $\mathbf{CA}(\mathrm{BEMB}, \mathrm{INPUT})$, both of which are order invariant in $\mathrm{INPUT}$, therefore the output of CMAB is order invariant in $\mathrm{INPUT}$.

As such, CMANPs are also context invariant as required. 

\textbf{Property: Target Equivariance.} A model $p_\theta$ is target
equivariant if for any choice of permutation function $\pi$, context datapoints $\{(x_i, y_i)\}_{i=1}^{N}$, and target datapoints $x_{N+1:N+M}$, 
\begin{align*}
p_\theta(y_{N+1:N+M} | x_{N+1:N+M} , x_{1:N}, y_{1:N}) = \\p_\theta(y_{\pi(N+1):\pi(N+M)}
| x_{\pi(N+1):\pi(N+M)}
, x_{1:N}, y_{1:N})
\end{align*}

\textbf{Proof Outline:} The vanilla variant of CMANPs makes predictions similar to that of LBANPs~\citep{feng2023latent} by retrieving information from a set of latent vectors via cross-attention and uses an MLP (Predictor). 
The architecture design ensures that the result is equivalent to making the predictions independently.
As such, CMANPs preserve target equivariance the same way LBANPs do.


\end{document}